%% file: neurips_2025.tex
\documentclass{article}

\PassOptionsToPackage{numbers, compress}{natbib}

\usepackage[preprint]{neurips_2025}

\usepackage{natbib}
\usepackage[utf8]{inputenc} 
\usepackage[T1]{fontenc}    
\usepackage{hyperref}       
\usepackage{url}            
\usepackage{booktabs}       
\usepackage{amsfonts}       
\usepackage{nicefrac}       
\usepackage{microtype}      
\usepackage{xcolor}         

\setcitestyle{square}
\setcitestyle{comma}
\usepackage{hyperref}
\hypersetup{
	colorlinks=true,
	citecolor=teal,
}
\usepackage{url}
\usepackage{booktabs}
\usepackage{microtype}      
\usepackage{multicol}
\usepackage{amsmath}
\usepackage{enumitem}
\usepackage{amssymb}
\numberwithin{equation}{section} 
\usepackage{wrapfig,lipsum,booktabs}
\usepackage{xcolor}
\usepackage{multirow}
\usepackage{float}
\usepackage{graphicx}
\usepackage{subcaption}
\usepackage{algpseudocode}  
\usepackage{algorithmicx,algorithm}
\usepackage{caption}
\usepackage{diagbox} 
\usepackage{colortbl}
\usepackage{xcolor}
\definecolor{rowcolor}{rgb}{0.9, 0.9, 0.9}
\usepackage{soul}
\usepackage{pifont}
\usepackage{enumitem}
\let\oldding\ding
\renewcommand{\ding}[2][1]{\scalebox{#1}{\oldding{#2}}}

\usepackage{cleveref}
\usepackage{marvosym}
\usepackage{stfloats}
\usepackage{tikz}
\usepackage{pgfplots}
\usepackage{tcolorbox}

\title{GuirlVG: Incentivize GUI Visual Grounding via Empirical Exploration on Reinforcement Learning}

\author{
    \textbf{Weitai Kang}$^{1}${ }\textbf{Bin Lei}$^{2}${ }\textbf{Gaowen Liu}$^{3}${ } \textbf{Caiwen Ding}$^2${ } \textbf{Yan Yan}$^{1}${ }\\
    $^1$University of Illinois Chicago { } \\
    $^2$University of Minnesota { }
    $^3$Cisco Research \\
}

\begin{document}
\maketitle
\input{sections/abs}

\input{sections/intro}
\input{sections/relate}
\input{sections/method}
\input{sections/exp}
\input{sections/con}
\clearpage
{\small
\bibliographystyle{plainnat}
\bibliography{reference}
}

\input{sections/append}
\end{document}

%% file: sections/abs.tex
\begin{abstract} 
Graphical user interface visual grounding (GUI-VG)—a core capability for GUI agents—has primarily relied on supervised fine-tuning (SFT) of multimodal large language models (MLLMs), demanding extensive data curation and significant training costs. 
However, as MLLMs continue to advance and even cover GUI domains during pretraining, the necessity of exhaustive SFT post-training becomes increasingly questionable.
Meanwhile, the recent successes of rule-based reinforcement fine-tuning (RFT) suggest a more efficient alternative.
However, despite its promise, the optimal manner of RFT for GUI-VG remains unexplored.
To bridge this gap, we introduce \textit{GuirlVG}, a reinforcement learning–based GUI-VG method built on a systematic empirical study and a novel stabilization technique.
Preliminarily, we find that naive application of RFT underperforms the SFT baseline, motivating a deeper exploration of RFT.
First, 
we decompose RFT into its core components and analyze the optimal formulation of each.
Second, 
as part of this exploration, we propose a novel Adversarial KL Factor that dynamically stabilizes training to mitigate reward over-optimization.
Third, 
we further explore the training configurations of RFT to enhance the effectiveness.
Extensive experiments show that \textit{GuirlVG}, with only 5.2K training samples, outperforms SFT methods trained on over 10M samples, achieving 
a \textbf{+7.7\%} improvement on ScreenSpot, a \textbf{+17.2\%} improvement on ScreenSpotPro and \textbf{91.9\%} accuracy on ScreenSpotV2.
\end{abstract}

\begin{figure*}[!h]
    \centering
    \vspace{-15pt}
    \includegraphics[width=0.9\textwidth]{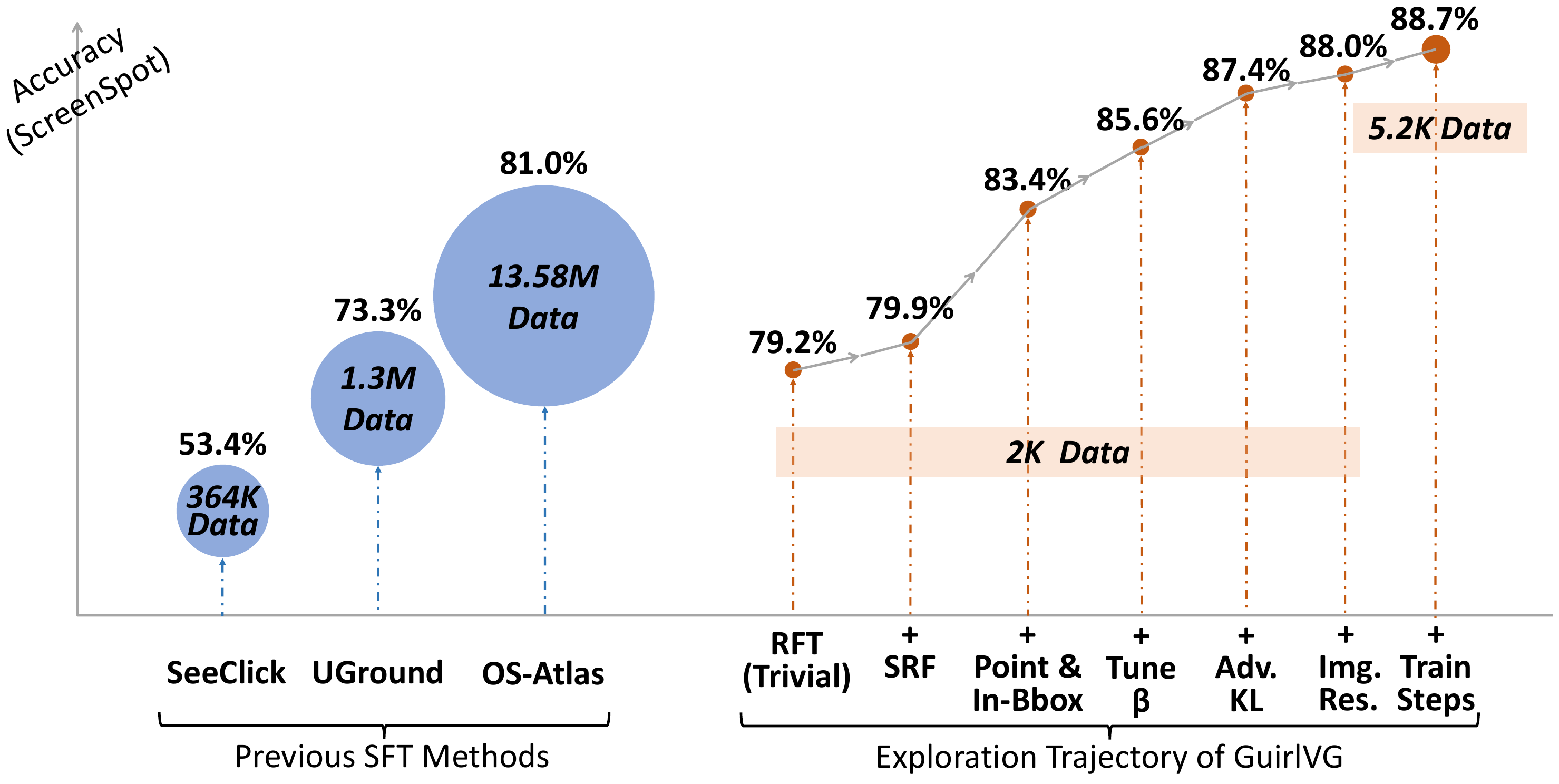}
    \caption{Step-by-step exploration of \textit{GuirlVG}. Starting from trivial RFT, we progressively add Soft Reward Function, In-Bbox reward with point prediction, $\beta$ tuning, our Adversarial KL Factor, image resolution prompting, and extended training. With only 5.2K data, \textit{GuirlVG} surpasses SFT methods trained on up to 13.58M data. Circle size reflects data scale used by each method.}
    \vspace{-15pt}
\end{figure*}

%% file: sections/intro.tex
\section{Introduction}

Graphical user interface (GUI) agents~\cite{gou2024navigating, lin2024showui, cheng2024seeclick, qin2025ui, xu2024aguvis, huang2025spiritsight, leigrounding, wu2024atlas, hong2024cogagent}, empowered by the rapid advancement of foundation models or multimodal large language models (MLLMs)~\cite{llava1.5, qwen2, qwen2.5}, are increasingly capable of perceiving and acting within digital environments via screenshots. A core capability underpinning such agents is GUI visual grounding (GUI-VG)---the task of localizing actionable elements in a screenshot conditioned on a textual instruction~\cite{gou2024navigating, cheng2024seeclick, leigrounding}. Recent efforts have primarily approached GUI-VG through post-training of MLLMs via supervised fine-tuning (SFT), a paradigm that demands large-scale domain-specific data curation and significant training resources~\cite{wu2024atlas, cheng2024seeclick, qin2025ui, gou2024navigating, leigrounding}. These advancements co-evolve with MLLM's capabilities, tailoring each generation of MLLMs to GUI-centric benchmarks.

However, this SFT paradigm raises critical concerns regarding efficiency. As MLLMs continue to improve in general perception and reasoning---with some already ingesting GUI-related data during pretraining~\cite{qwen2, qwen2.5}---the necessity of extensive post-training becomes increasingly questionable. Given the persistent training cost incurred with each new MLLM generation, a fundamental question arises: \textit{Does exhaustive SFT remain the most effective post-training strategy?}

Meanwhile, the success of rule-based reinforcement fine-tuning (RFT) by Group Relative Policy Optimization (GRPO)~\cite{grpo} in DeepSeek-R1~\cite{guo2025deepseek} inspires new directions. This method has recently been extended to different domains~\cite{liu2025visual, wang2025visualprm, peng2025lmm, chen2025r1v} with notable improvements, offering a promising path toward a more efficient post-training.
Despite these advances, no prior work has systematically studied RFT for GUI-VG. In fact, our results even reveal that naive application of RFT to GUI-VG under fair experimental settings underperforms the SFT baseline, prompting a critical question: \textit{What is the optimal formulation of RFT objectives for GUI visual grounding?}

In this paper, we introduce \textit{GuirlVG}, a reinforcement learning–based method for GUI visual grounding, built upon a comprehensive empirical study of RFT and a novel stabilization technique toward GRPO.
\raisebox{-1.1pt}{\ding[1.1]{182\relax}}
We begin by deconstructing GRPO into its core components—format reward, accuracy reward, and KL penalty—and systematically ablate each component to derive an optimal configuration.
\raisebox{-1.1pt}{\ding[1.1]{183\relax}}
To further address over-optimization caused by reward functions, we introduce a novel Adversarial KL Factor, which dynamically scales the KL penalty based on rewards to stabilize the learning process.
\raisebox{-1.1pt}{\ding[1.1]{184\relax}}
Additionally, we explore a wide range of training setups, including hyperparameter tuning, LoRA enablement, and prompt engineering, to uncover best practices for effective RFT on GUI-VG.
\raisebox{-1.1pt}{\ding[1.1]{185\relax}}
Finally, we conduct extensive experiments on ScreenSpot~\cite{cheng2024seeclick}, ScreenSpotV2~\cite{wu2024atlas}, and ScreenSpotPro~\cite{li2025screenspot}, demonstrating that \textit{GuirlVG} achieves state-of-the-art results using as few as 2K$\sim$5.2K training examples. Compared to prior SFT baselines trained on hundreds of thousands to over ten million data, our method achieves superior accuracy with up to \textbf{+17.2\%} absolute gains on ScreenSpotPro, highlighting the data efficiency and strong effectiveness of \textit{GuirlVG}.

%% file: sections/relate.tex
\section{Background}

\subsection{Related Work}
\paragraph{GUI Visual Grounding.} Enabling AI agents to perform automation within Graphical User Interfaces (GUIs) has gained increasing traction, as it allows models to operate directly in software environments and alleviate human workload. This capability requires both high-level task planning for different actions and the accurate grounding of GUI elements where actions are applied.
Earlier work~\cite{koh2024visualwebarena, zhou2023webarena, cao2024spider2} typically leverages HTML structures or a11y accessibility trees to support grounding. However, many commercial applications are closed-source, limiting access to such internal resources. This constraint has led to a growing interest in vision-based agents~\cite{gou2024navigating, lin2024showui, cheng2024seeclick, qin2025ui, xu2024aguvis, huang2025spiritsight, leigrounding, wu2024atlas, hong2024cogagent}, which operate solely on screenshots as visual observations.
Consequently, the visual grounding ability~\cite{intent3d, segvg, attbalance, actress, transvg, kang2025robin3dimproving3dlarge} to localize actionable elements based on the screenshot—GUI Visual Grounding (GUI-VG)~\cite{gou2024navigating, cheng2024seeclick, leigrounding}—has become the main bottleneck for these methods~\cite{cheng2024seeclick, gou2024navigating, leigrounding}.
To address this, SeeClick~\cite{cheng2024seeclick} introduces a large-scale pretraining pipeline for GUI-VG and proposes an automated method to generate training data. Similarly, UGround~\cite{gou2024navigating} utilizes synthesized web-based data to support grounding training, and AGG~\cite{leigrounding} builds a dedicated engine to collect extensive GUI images with annotations. OS-Atlas~\cite{wu2024atlas} further expands grounding data across multiple operating systems.
UI-TARs~\cite{qin2025ui} combines GUI-centric pretraining with task-conditioned fine-tuning to improve alignment between perception and reasoning. Despite the variety in their data construction, these methods commonly adopt the supervised fine-tuning (SFT) paradigm, which relies heavily on large volumes of high-quality labeled training data.

\paragraph{Reinforcement Fine-Tuning.} 
Rule-based Reinforcement Fine-Tuning (RFT) with Group Relative Policy Optimization (GRPO)~\cite{grpo} has recently demonstrated effectiveness across domains such as mathematical reasoning~\cite{shao2024deepseekmath}, code generation~\cite{code-r1}, and logical inference~\cite{RAGEN}, with a notable example being DeepSeek-R1~\cite{guo2025deepseek}. 
Unlike supervised fine-tuning (SFT), which enforces token-level supervision strictly corresponding to the answer, RFT encourages models to freely explore their reasoning process and provides supervision only at the level of the final output. 
This more flexible objective incentivizes stronger reasoning capabilities~\cite{guo2025deepseek}.
Furthermore, in the RFT algorithm---GRPO, task-specific rule-based reward functions are designed to provide supervision signals that are automatically verifiable. 
This eliminates the need for training a separate critic model~\cite{schulman2017proximal, ouyang2022training} or relying on human feedback~\cite{kaufmann2023survey}, thereby mitigating the risk of reward hacking~\cite{weng2024rewardhacking} and making RFT an effective alternative to SFT.
Despite its potential, there remains a lack of empirical studies investigating RFT in the context of GUI-VG, where unique challenges such as diverse layouts and high-resolution visual inputs require models to integrate fine-grained spatial understanding with broader contextual reasoning.

\subsection{Preliminaries}

\paragraph{Group Relative Policy Optimization (GRPO).} 
Given a task input which additionally specifies a particular response format in the prompt, i.e. instructing the model to reason within $<think> <think>$ tags and answer within $<answer> </answer>$ tags, the model generates a group of $N$ candidate responses $\{o_1, o_2, \dots, o_N\}$. 
Each candidate is evaluated using a rule-based reward function, yielding rewards $\{r_1, r_2, \dots, r_N\}$. 
For each response $o_i$,
this rule-based reward function scores two rewards: a format reward, $r^f_i$, which assesses whether the response adheres to the instructed tag structure, and an accuracy reward, $r^a_i$, which evaluates the correctness of the response, such as classification accuracy~\cite{chen2025r1v} or intersection-over-union (IoU) in detection tasks~\cite{huang2025vision, liu2025visual}.
The total reward for is computed as
\begin{equation}
r_i = r^f_i + r^a_i.
\end{equation}
The relative reward (also referred to as the advantage $A_i$) of the $i$-th candidate is computed by normalizing the rewards within the group of candidate responses:
\begin{equation}
A_i = \frac{r_i - \text{Mean}(\{r_1, r_2, \dots, r_N\})}{\text{Std}(\{r_1, r_2, \dots, r_N\})},
\end{equation}
where $\text{Mean}(\cdot)$ and $\text{Std}(\cdot)$ denote the mean and standard deviation, respectively.
To stabilize training, GRPO additionally constrains model update by minimizing the KL divergence between the current model and a reference model (typically the original model). 
Thus, the objective $J_i$ to maximize for each candidate $o_i$ becomes
\begin{equation}
J_i = A_i - \beta \, \mathbb{D}_{\text{KL}}(o_i \, \| \, o_i^{\text{orig}}),
\end{equation}
where $\beta$ is a hyperparameter controlling the KL penalty strength, and $o_i^{\text{orig}}$ is the corresponding response from the reference model. We omit details, such as clipping, averaging, etc.

\paragraph{Implementation.}
Unless specified otherwise, we fine-tune Qwen2.5-VL~\cite{qwen2.5} using LoRA~\cite{hu2022lora} with a rank of 64 and an alpha of 128, while keeping the vision module frozen. 
Training data are randomly sampled from ShowUI~\cite{lin2024showui}, which crawls visually rich website data and augments desktop data from OmniAct~\cite{kapoor2024omniact} using GPT-4o~\cite{hurst2024gpt}.
The group size of candidate responses, $N$, is set to 6, and the batch size is set to 4. 
The KL divergence coefficient ($\beta$) is set to 0.04 by default.
The learning rate is set to $1\times10^{-5}$, with two training epochs, AdamW optimizer, and a linear decay schedule.
We use 6$\times$NVIDIA A100-80G GPUs for training.
For the SFT baseline, we adopt LLaMA Factory~\cite{zheng2024llamafactory} with the same training configurations for a fair comparison.
For the efficiency and fairness of experiments, we report performances at step 500 for both RFT and SFT, where convergence is typically observed.
Training beyond 500 steps yields only marginal improvements, with our final version reaching peak performance around step 1,300.
Accordingly, our final version is only trained on 5,200 samples.

\paragraph{Evaluation Suite.}\label{evaluate}
We evaluate on three widely-used GUI-VG benchmarks across different platforms: ScreenSpot~\cite{cheng2024seeclick}, ScreenSpot v2~\cite{wu2024atlas}, and ScreenSpot-Pro~\cite{li2025screenspot}. 
ScreenSpot evaluates GUI grounding capabilities across mobile, desktop, and web environments, while ScreenSpot v2 improves evaluation reliability by correcting annotation errors. 
ScreenSpot-Pro focuses on high-resolution professional scenarios, featuring expert-annotated tasks spanning 23 applications, five industries, and three operating systems.
All benchmarks report the accuracy of whether the predicted point coordinate falls inside the ground truth bounding box of the corresponding element in the screenshot.

%% file: sections/method.tex
\section{Methodology}

\subsection{Can trivial adoption of RFT beats SFT?}\label{rft_vs_sft}
We begin by comparing the SFT baseline with a trivial adoption of RFT for GUI-VG.
Specifically, we adopt the commonly used implementation from \citet{openr1, shen2025vlm}, using the following prompt for a given description of the target element:

\begin{wraptable}{r}{0.6\textwidth}
\centering
\caption{Comparison of zero-shot, SFT, and trivial RFT on ScreenSpot (Qwen2.5-VL, 500 training steps).}
\label{trivial}
\begin{tabular}{lccc}
\toprule
Method & Backbone & Step & Avg. Acc (\%) \\
\midrule
Zero-Shot & Qwen2.5-VL & 500 & 72.6 \\
SFT       & Qwen2.5-VL & 500 & 82.6 \\
RFT (trivial) & Qwen2.5-VL & 500 & 79.2 \\
\bottomrule
\end{tabular}
\end{wraptable}

\textit{Please provide the bounding box coordinates [x1, y1, x2, y2] of a specific element based on this sentence: <description>. First, think through the reasoning process within <think> </think> tags. Then, output the bounding box coordinates in JSON format within <answer> </answer> tags.}

For the format reward, a value of 1 is assigned if the output exactly matches the pattern ``<think>...</think>...<answer>...</answer>'', and 0 otherwise.
The accuracy reward assigns 1 if a bounding box (bbox) array enclosed in a square bracket is detected and the IoU between the predicted and ground-truth bboxes exceeds 0.5, and 0 otherwise.
During inference, the center of the predicted bbox is used as final prediction.
Due to space limitations, we provide the detailed pseudo-code of RFT (trivial), along with the implementation details of the SFT baseline and the zero-shot setup for Qwen2.5-VL, in \cref{appendix_rft_vs_sft}.
As shown in \cref{trivial}, both SFT and trivial RFT lead to improvements over the zero-shot baseline, but RFT (trivial) does not outperform SFT.
\begin{figure*}[h]
    \centering
    \includegraphics[width=1\textwidth]{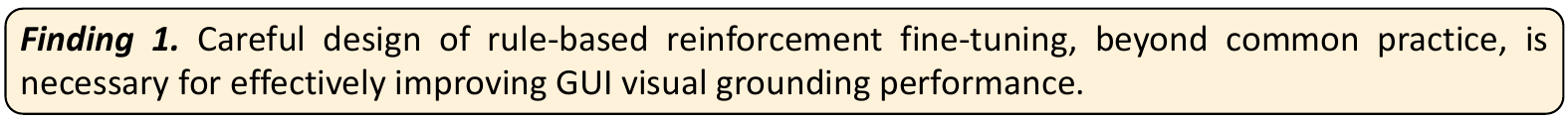}
\end{figure*}

\subsection{How to design reward functions in GRPO?}\label{reward_grpo}

As defined earlier, the default format reward enforces exact tag matching, while the accuracy reward relies strict JSON-style output consistent with the model’s pretraining. The model is sharply penalized (rewarded 0) if any part of the expected structure is missing—such as an omitted \texttt{</answer>} tag—or it has a minor style deviation, e.g., outputting coordinates as a tuple instead of a JSON list. This rigid design introduces training noise and instability, even when the model successfully performs reasoning and answering.

\begin{wraptable}{r}{0.6\textwidth}
\centering
\caption{Compare the default reward function and our SRF on ScreenSpot (Qwen2.5-VL, 500 training steps).}
\label{soft_vs_strict}
\begin{tabular}{lccc}
\toprule
Method & Backbone & Step & Avg. Acc (\%) \\
\midrule
Default     & Qwen2.5-VL & 500 & 79.2 \\
SRF (Ours) & Qwen2.5-VL & 500 & 79.9 \\
\bottomrule
\end{tabular}
\end{wraptable}

To address this, we propose the Soft Reward Function (SRF), which provides partial credit to the presence of each tag and relaxes output style. Specifically, SRF removes the JSON requirement from the prompt.
For the format reward, SRF assigns +0.5 for each of \texttt{<think>} and \texttt{</think>}, +1/3 for each of \texttt{<answer>} and \texttt{</answer>}, and +1/3 if the content inside the answer tags contains the correct number of coordinates. The total score is normalized to [0, 1].
For the accuracy reward, SRF ignores style and simply extracts numeric values present in the output.
Detailed prompts and pseudo-code are provided in \cref{appendix_reward_grpo} due to space constraints.
As shown in \cref{soft_vs_strict}, SRF provides +0.7\% improvement over the default reward function.

\begin{figure*}[h]
    \centering
    \includegraphics[width=1\textwidth]{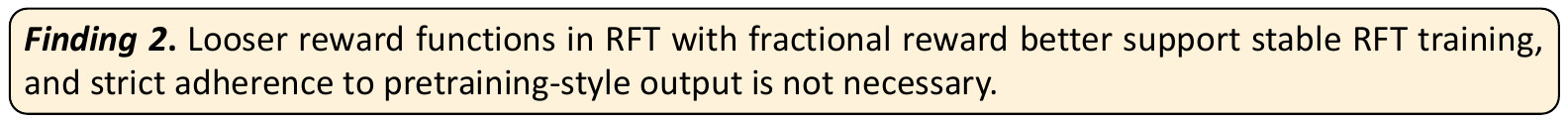}
\end{figure*}

\subsection{How to design model prediction format along with its accuracy reward function?}

The goal of GUI visual grounding is to predict a point that falls within the target element to enable the downstream action. To support this functionality, the most direct design is to predict a point and assign a binary reward based on whether it lies within the ground-truth bounding box~\cite{shen2025vlm} (In-Bbox).
Alternatively, one can define reward based on a distance threshold $k$~\cite{liu2025seg}, where the point prediction is rewarded with 1 if it falls within $k$ pixels of the target center (denoted as Distance@$k$).
Another option is to output a bounding box and derive a point prediction from its center, evaluating it via IoU with the ground truth. This can be used as a continuous reward or a thresholded one (e.g., IoU@0.5 gives a reward of 1 if IoU $>$ 0.5, and 0 otherwise, as in our default format).

\begin{table*}[h]\centering
\centering
\caption{Comparison of different prediction formats and accuracy reward functions under SRF on ScreenSpot (Qwen2.5-VL, 500 training steps).}
\label{prediction_format}
\begin{tabular}{ccccc}
\toprule
Prediction & Reward & Backbone & Step & Avg. Acc (\%) \\
\midrule
Bbox & IoU@0.5     & Qwen2.5-VL & 500 & 79.9 \\
Bbox & IoU        & Qwen2.5-VL & 500 & 81.6 \\
Point & Distance@80 & Qwen2.5-VL & 500 & 82.7 \\
Point & In-Bbox     & Qwen2.5-VL & 500 & 83.4 \\
\bottomrule
\end{tabular}
\end{table*}

Building on our Soft Reward Function,
we evaluate four configurations. The threshold of 80 for Distance@$k$ is empirically selected for best performance. As shown in \cref{prediction_format}, Point prediction with In-Bbox performs best.
\begin{figure*}[h]
    \centering
    \includegraphics[width=1\textwidth]{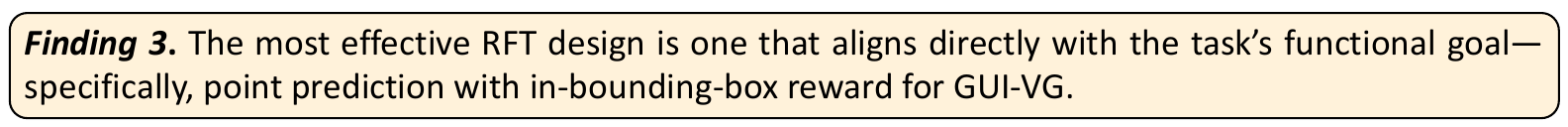}
\end{figure*}

\subsection{How to balance the KL Penalty in GRPO?}

In GRPO, the KL penalty term enforces the current model to stay close to the original model, mitigating reward-driven over-optimization~\cite{grpo, guo2025deepseek}. The hyperparameter \( \beta \) plays a critical role in determining the strength of this regularization. In our experiments, we observed that model performance is highly sensitive to this parameter.

We first empirically explore the effect of different values for \( \beta \), then introduce a novel strategy we call \textit{Adversarial KL Factor}, which dynamically scales the KL penalty based on reward strength. The intuition is that high-reward responses are more likely to cause over-optimization in GRPO. However, the KL penalty with the original model does not necessarily increase proportionally, especially when the original model itself assigns high probability to such responses. Therefore, a static KL term may fail to counterbalance the effect of reward.
To address this, we define the \textit{Adversarial KL Factor} as the ratio of the reward to its theoretical maximum \( m \), and use it as a multiplicative modifier to \( \beta \) to scale the KL penalty proportionally.
This dynamic formulation ensures that as reward increases, the regularization also strengthens adaptively. The modified GRPO objective is:
\begin{equation}
J_i = A_i - \alpha_i \beta \, \mathbb{D}_{\text{KL}}(o_i \, \| \, o_i^{\text{orig}}),\quad
A_i = \frac{r_i - \text{Mean}(\{r_1, r_2, \dots, r_N\})}{\text{Std}(\{r_1, r_2, \dots, r_N\})},\quad
\alpha_i = \frac{r_i}{m},
\end{equation}
where \( m = 2 \) is the maximum possible reward under our setup.

\begin{table*}[!h]\centering
\centering
\caption{Comparison of different KL settings under SRF, point prediction, and In-Bbox reward on ScreenSpot (Qwen2.5-VL, 500 training steps).}
\label{adversarial}
\begin{tabular}{ccccc}
\toprule
Adversarial & \( \beta \) & Backbone & Step & Avg. Acc (\%) \\
\midrule
\ding{55} & 4e-2 & Qwen2.5-VL & 500 & 83.4 \\
\ding{55} & 0 & Qwen2.5-VL & 500 & 84.7 \\
\ding{55} & 1e-4 & Qwen2.5-VL & 500 & 85.6 \\
\ding{51} & 1e-4 & Qwen2.5-VL & 500 & 87.4 \\
\ding{51} & 1e-6 & Qwen2.5-VL & 500 & 77.5 \\
\bottomrule
\end{tabular}
\end{table*}

Results are shown in \cref{adversarial}. Simply tuning \( \beta \) provides clear performance improvements, demonstrating the importance of empirically calibrating the KL penalty. Notably, our \textit{Adversarial KL Factor} strategy (row 4) achieves a substantial +1.8\% gain over the best \( \beta \) baseline (row 3), validating the advantage of dynamically adjusting KL strength in response to reward magnitude. Row 5 further indicates that setting \( \beta \) too small results in degraded performance.

\begin{figure*}[h]
    \centering
    \includegraphics[width=1\textwidth]{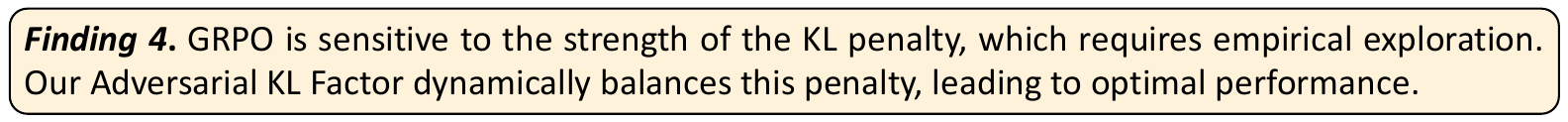}
\end{figure*}

\subsection{Should we fully fine-tune the model or use LoRA?}

We further investigate the impact of fine-tuning strategies by comparing full model fine-tuning (Full-FT) with LoRA~\cite{lora} fine-tuning (LoRA-FT) applied to the LLM component. 
In practice, we observe that full fine-tuning tends to destabilize training unless a much smaller learning rate is used. Therefore, we reduce the learning rate for full fine-tuning to \( 1 \times 10^{-6} \), while keeping other hyperparameters consistent. We also report the training time per iteration using 6$\times$A6000 GPUs.

\begin{table*}[!h]\centering
\centering
\caption{Comparison of Full-FT and LoRA-FT under SRF, point prediction, In-Bbox reward, $\beta = 1 \times 10^{-4}$, and Adversarial KL Factor on ScreenSpot (Qwen2.5-VL, 500 training steps). Training time is reported per iteration over 6$\times$A6000 GPUs.}
\label{LoRA}
\begin{tabular}{lcccc}
\toprule
Config & Backbone & Step & Time & Avg. Acc (\%) \\
\midrule
Full-FT & Qwen2.5-VL & 500 & 749.4 s & 87.5 \\
LoRA-FT & Qwen2.5-VL & 500 & 28.4 s & 87.4 \\
\bottomrule
\end{tabular}
\end{table*}
As shown in \cref{LoRA}, Full-FT requires over 25 times more training time per iteration compared to LoRA-FT, while yielding only a marginal improvement of +0.1\%. Given this modest performance gain relative to the substantial increase in computational cost, we adopt LoRA-FT as a more efficient strategy for GUI-VG reinforcement fine-tuning in our study.

\begin{figure*}[!h]
    \centering
    \includegraphics[width=1\textwidth]{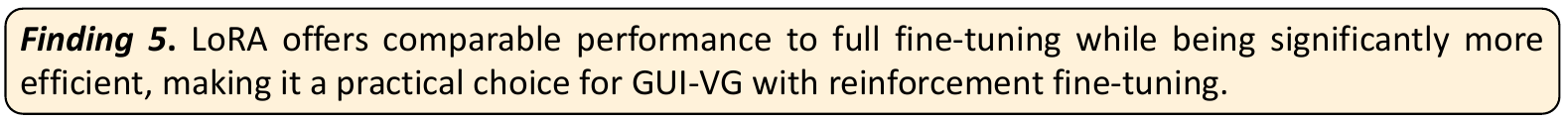}
\end{figure*}

\subsection{How to decide the group size and batch size in GRPO?}

The hyperparameters group size and batch size play critical roles in GRPO~\cite{grpo}. Specifically, group size affects the normalization of advantage estimates, while batch size determines how each sample contributes to the final objective function. Therefore, it is necessary to empirically examine how different configurations of these two hyperparameters impact the final performance.

As shown in \cref{size}, the configuration with group size 6 and batch size 4 achieves the highest accuracy, which is our default setting. Interestingly, increasing the group size from 6 to 8 leads to a substantial performance drop, even though larger groups theoretically provide better baseline estimates for advantage in GRPO to serve as a more stable substitute for the critic model in PPO~\cite{schulman2017proximal}. This counterintuitive result suggests that RFT is sensitive to seemingly minor changes in implementation details and highlights the need for systematic validation of hyperparameter choices.

\begin{table*}[!h]\centering
\centering
\caption{Effect of group size and batch size under SRF, point prediction, In-Bbox reward, $\beta = 1 \times 10^{-4}$, Adversarial KL Factor and LoRA on ScreenSpot (Qwen2.5-VL, 500 training steps).}
\label{size}
\begin{tabular}{ccccc}
\toprule
Group & Batch & Backbone & Step & Avg. Acc (\%) \\
\midrule
6 & 1 & Qwen2.5-VL & 500 & 86.5 \\
6 & 4 & Qwen2.5-VL & 500 & 87.4 \\
8 & 4 & Qwen2.5-VL & 500 & 83.9 \\
\bottomrule
\end{tabular}
\end{table*}

\begin{figure*}[!h]
    \centering
    \includegraphics[width=1\textwidth]{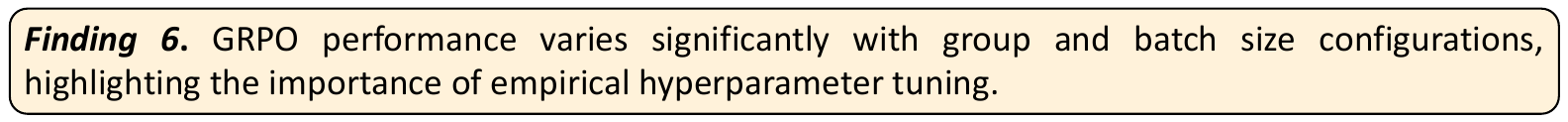}
\end{figure*}

\subsection{How to involve image resolution information in the prompt?}
Prompting image resolution may provide additionally helpful context, especially for high-resolution GUI screenshots. We explore when such information should be incorporated into the prompt. Specifically, we compare three strategies: (1) never provide resolution; (2) provide resolution during both training and testing; (3) provide resolution only at test time. When resolution is included, we prepend the prompt with \textit{"The screenshot resolution is \{width\}$\times$\{height\}."}

\begin{table*}[!h]\centering
\centering
\caption{Effect of image resolution in the prompt under SRF, point prediction, In-Bbox reward, LoRA, $group size$ = 6, and $batchsize$ = 4 on ScreenSpot (Qwen2.5-VL, 500 training steps).}
\label{img_res}
\begin{tabular}{ccccc}
\toprule
Train & Test & Backbone & Step & Avg. Acc (\%) \\
\midrule
\ding{51} & \ding{51} & Qwen2.5-VL & 500 & 83.7 \\
\ding{55} & \ding{55} & Qwen2.5-VL & 500 & 87.4 \\
\ding{55} & \ding{51} & Qwen2.5-VL & 500 & 88.0 \\
\bottomrule
\end{tabular}
\end{table*}
As shown in \cref{img_res}, the highest accuracy is achieved when resolution information is excluded during training but added at test time. We hypothesize that withholding resolution during training may challenge the model to learn a better spatial reasoning ability. At test time, the additional resolution context then serves as a useful signal to refine predictions.

\begin{figure*}[!h]
    \centering
    \includegraphics[width=1\textwidth]{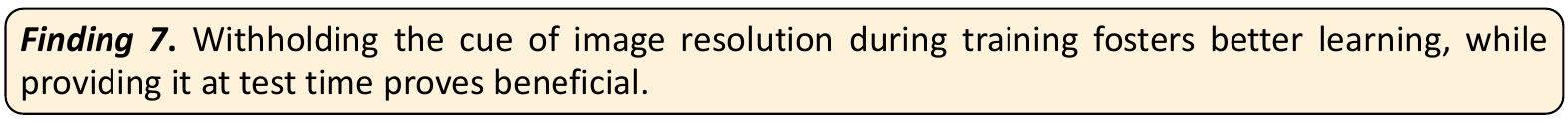}
\end{figure*}

\subsection{Final Design Choices for RFT on GUI-VG}
Based on the studies above, we finalize a set of design choices for an effective and efficient RFT pipeline for GUI visual grounding under GRPO. We propose the Soft Reward Function (SRF) to provide partial credit for format compliance while relaxing output constraints. For the prediction format, we use direct point prediction with the In-Bbox binary reward. To stabilize training, we introduce the Adversarial KL Factor with a coefficient of $\beta = 1 \times 10^{-4}$. We employ LoRA for efficient fine-tuning and set the group size to 6 and batch size to 4. Image resolution information is withheld during training and added only at inference. We train 1,300 steps for our final version.

%% file: sections/exp.tex
\section{Comparison with preivous methods}
We compare our final RFT method against prior approaches across three GUI-VG benchmarks introduced in \cref{evaluate}: ScreenSpot~\cite{cheng2024seeclick}, ScreenSpot v2~\cite{wu2024atlas}, and ScreenSpot-Pro~\cite{li2025screenspot}.

\begin{table*}[h]\centering
\caption{Comparison of various models on ScreenSpot. The optimal result is \textbf{bolded}. ``Size'' refers to model size. ``\#Train'' refers to training samples.} 
\vspace{-6pt}
\resizebox{0.9\textwidth}{!}{%
\begin{tabular}{llcrrrrrrrr}\toprule
\multirow{2}{*}{Method} &\multirow{2}{*}{Size} &\multirow{2}{*}{\#Train} &\multicolumn{2}{c}{{Mobile}} &\multicolumn{2}{c}{{Desktop}} &\multicolumn{2}{c}{{Web}} &\multirow{2}{*}{{Avg.}} \\
& & &Text &Icon &Text &Icon &Text &Icon & \\\midrule
Fuyu~\cite{fuyu-8b} & 8B &--&41.0 &1.3 &33.0 &3.6 &33.9 &4.4 & \cellcolor{yellow!10} 19.5 \\
CogAgent~\cite{cogagent} &18B & \cellcolor{yellow!10}{400K} &67.0 &24.0 &74.2 &20.0 &{70.4} &28.6 & \cellcolor{yellow!10} 47.4 \\
SeeClick~\cite{cheng2024seeclick} & 9.6B & \cellcolor{yellow!10}{364K} & {78.0} &52.0 &72.2 &30.0 &55.7 &32.5 & \cellcolor{yellow!10}53.4 \\
AGG~\cite{leigrounding} & 0.4B & \cellcolor{yellow!10}{35M} & 86.1 & 62.8 & 81.8 & 46.2 & 74.2 & 48.4 & \cellcolor{yellow!10}66.6 \\
OmniParser~\cite{omniparser} & \small * & -- & 93.9 & 57.0 & 91.3 & 63.6 & 81.3 & 51.0 &  \cellcolor{yellow!10}73.0\\
UGround~\cite{gou2024navigating} & 7B & \cellcolor{yellow!10}{1.3M} & 82.8 & 60.3 & 82.5 & 63.6 & 80.4 & 70.4 &  \cellcolor{yellow!10}73.3\\
ShowUI-G~\cite{lin2024showui} & 2B & \cellcolor{yellow!10}{119K} & 91.6 & 69.0 & 81.8 & 59.0 & 83.0 & 65.5 &  \cellcolor{yellow!10}74.9 \\
ShowUI~\cite{lin2024showui} & 2B & \cellcolor{yellow!10}{256K} & 92.3 & 75.5 & 76.3 & 61.1 & 81.7 & 63.6 &  \cellcolor{yellow!10}75.1 \\
OS-Atlas~\cite{wu2024atlas} & 4B & \cellcolor{yellow!10}{13.58M} & 85.7 & 58.5 & 72.2 & 45.7 & 82.6 & 63.1 &  \cellcolor{yellow!10}68.0 \\
OS-Atlas~\cite{wu2024atlas} & 7B & \cellcolor{yellow!10}{13.58M} & 93.0 & 72.9 & 91.8 & 62.9 & 90.9 & 74.3 &  \cellcolor{yellow!10}81.0 \\
\midrule
GuirlVG & 7B & \cellcolor{yellow!10}{2K}  & 96.3  & 86.0 & 93.3 & 77.1 & 91.7 & 83.5  & \cellcolor{yellow!10}88.0 \\
GuirlVG & 7B & \cellcolor{yellow!10}{5.2K}  & 96.0  & 84.7 & 92.8 & 80.0 & 92.6 & 85.9  &  \bf \cellcolor{yellow!10} 88.7 \\
\bottomrule
\end{tabular}
}
\label{tab:screenspot}
\end{table*}

 Results on the ScreenSpot benchmark are shown in \cref{tab:screenspot}. Our method substantially outperforms previous methods that rely on supervised fine-tuning (SFT), despite using significantly fewer training samples. Specifically, while prior SFT methods are trained on hundreds of thousands to over ten million examples, our RFT method achieves superior performance with just 2K training samples. For example, we outperform OS-Atlas—\textit{which uses 6.79K times more data}—by \textbf{+7.0\%} in accuracy, highlighting the efficiency and effectiveness of RFT as a post-training strategy. When increasing training to 1300 steps using 5.2K training samples, our method achieves further improvements, outperforming OS-Atlas by \textbf{+7.7\%}. Notably, on the Mobile-Icon subset, our method exceeds OS-Atlas by \textbf{+11.8\%}, despite our training data containing no mobile-specific samples. This suggests that RFT enhances out-of-domain reasoning capabilities, aligning with the claim from \citet{chu2025sft} that ``SFT memorizes, RL generalizes.''

\begin{table*}[h]\centering
\caption{Comparison of various models on ScreenSpot v2. The optimal result is \textbf{bolded}. ``Size'' refers to model size. ``\#Train'' refers to training samples.} 
\resizebox{0.9\textwidth}{!}{%
\begin{tabular}{llcrrrrrrrr}
\toprule
\multirow{2}{*}{Method} &\multirow{2}{*}{Size} &\multirow{2}{*}{\#Train} &\multicolumn{2}{c}{{Mobile}} &\multicolumn{2}{c}{{Desktop}} &\multicolumn{2}{c}{{Web}} &\multirow{2}{*}{{Avg.}} \\
& & &Text &Icon &Text &Icon &Text &Icon & \\\midrule
SeeClick~\cite{cheng2024seeclick} & 9.6B & \cellcolor{yellow!10}{364K} & 78.4 & 50.7 & 70.1 & 29.3 & 55.2 & 32.5 & \cellcolor{yellow!10}55.1 \\
OS-Atlas~\cite{wu2024atlas} & 4B & \cellcolor{yellow!10}{13.58M} & 87.2 & 59.7 & 72.7 & 46.4 & 85.9 & 63.1 & \cellcolor{yellow!10}71.9 \\
OS-Atlas~\cite{wu2024atlas} & 7B & \cellcolor{yellow!10}{13.58M} & 95.2 & 75.8 & 90.7 & 63.6 & 90.6 & 77.3 & \cellcolor{yellow!10}84.1 \\
\midrule
GuirlVG & 7B & \cellcolor{yellow!10}{2K}  & 99.3  & 89.6 & 94.8 & 72.9 & 95.7 & 83.3  & \cellcolor{yellow!10}90.9 \\
GuirlVG & 7B & \cellcolor{yellow!10}{5.2K}  & 98.3  & 89.6 & 94.3 & 80.7 & 95.7 & 86.2  &  \bf \cellcolor{yellow!10} 91.9 \\
\bottomrule
\end{tabular}
}
\label{tab:screenspotv2}
\end{table*}

Results on ScreenSpot v2 (\cref{tab:screenspotv2}) mirror the trends observed on ScreenSpot. With only 2K training examples, our method surpasses all previous methods, and with 5.2K examples, it reaches a new state-of-the-art of \textbf{91.9\%} average accuracy—\textbf{+7.8\%} higher than OS-Atlas (7B). Performance gains are consistent across all subdomains, reaffirming the generalization strength of our RFT pipeline.

\begin{table}[h]
    \centering
    \vspace{-6pt}
    \caption{Comparison of various models on ScreenSpot-Pro. The optimal result is \textbf{bolded}.} 
    \resizebox{\textwidth}{!}{%
        \begin{tabular}{lccccccccccccc}
        \toprule
        \multirow{2}{*}{{Model}} & \multicolumn{2}{c}{{Development}} & \multicolumn{2}{c}{{Creative}} & \multicolumn{2}{c}{{CAD}} & \multicolumn{2}{c}{{Scientific}} & \multicolumn{2}{c}{{Office}} & \multicolumn{2}{c}{{OS}} & \multirow{2}{*}{{Avg}} \\
        \cmidrule(lr){2-3} \cmidrule(lr){4-5} \cmidrule(lr){6-7} \cmidrule(lr){8-9} \cmidrule(lr){10-11} \cmidrule(lr){12-13} 
         & {Text} & {Icon} & {Text} & {Icon} & {Text} & {Icon} & {Text} & {Icon} & {Text} & {Icon} & {Text} & {Icon} &  \\
        \midrule
        SeeClick~\cite{cheng2024seeclick}         & $0.6$  & $0.0$  & $1.0$  & $0.0$  & $2.5$  & $0.0$  & $3.5$  & $0.0$  & $1.1$  & $0.0$  & $2.8$  & $0.0$  & \cellcolor{yellow!10}1.1 \\
        OS-Atlas-4B~\cite{wu2024atlas}      & $7.1$  & $0.0$  & $3.0$  & $1.4$  & $2.0$  & $0.0$  & $9.0$  & $5.5$  & $5.1$  & $3.8$  & $5.6$  & $0.0$  & \cellcolor{yellow!10}3.7 \\
        ShowUI-2B~\cite{lin2024showui} & $16.9$ & $1.4$  & $9.1$  & $0.0$  & $2.5$  & $0.0$  & $13.2$ & $7.3$  & $15.3$ & $7.5$  & $10.3$ & $2.2$  & \cellcolor{yellow!10}7.7 \\
        CogAgent-18B~\cite{cogagent} & $14.9$ & $0.7$  & $9.6$  & $0.0$  & $7.1$  & $3.1$  & $22.2$ & $1.8$  & $13.0$ & $0.0$  & $5.6$  & $0.0$  & \cellcolor{yellow!10}7.7 \\
        Aria-GUI~\cite{yang2024aria} & $16.2$ & $0.0$  & $23.7$ & $2.1$  & $7.6$  & $1.6$  & $27.1$ & $6.4$  & $20.3$ & $1.9$  & 4.7  & $0.0$  & \cellcolor{yellow!10}11.3 \\
        UGround-7B~\cite{gou2024navigating} & {26.6} & $2.1$  & $27.3$ & $2.8$  & {14.2} & $1.6$  & $31.9$ & $2.7$  & $31.6$ & $11.3$ & {17.8} & $0.0$  & \cellcolor{yellow!10}16.5 \\
        OS-Atlas-7B~\cite{wu2024atlas} & {33.1} & $1.4$  & $28.8$ & $2.8$  & $12.2$ & {4.7}  & {37.5} & $7.3$  & $33.9$ & $5.7$  & {27.1} & {4.5}  & \cellcolor{yellow!10}{18.9} \\
        \midrule
        GuirlVG-2K-7B & {57.8} & {9.0} & {38.9} & {10.5} & {26.9} & {7.8} & {44.4} & {14.5} & {57.1} & {22.6} & {39.3} & {14.6} & \cellcolor{yellow!10}{31.6} \\
        GuirlVG-5.2K-7B & {64.9} & {7.6} & {42.9} & {11.2} & {28.9} & {9.4} & {63.9} & {16.4} & {63.8} & {26.4} & {43.9} & {13.5} & \cellcolor{yellow!10}\textbf{36.1} \\
        \bottomrule
    \end{tabular}
   }
    \label{tab:screenspotpro}
\end{table}

Finally, results on ScreenSpot-Pro (\cref{tab:screenspotpro}) demonstrate the strong generalization of our method to high-resolution, professional GUIs. With only 2K training examples, our approach already outperforms all prior methods by a large margin, achieving \textbf{31.6\%} average accuracy—surpassing the best SFT baseline OS-Atlas (7B) by \textbf{+12.7\%}. Scaling up to 5.2K examples further boosts performance to \textbf{36.1\%}, an absolute gain of \textbf{+17.2\%} over OS-Atlas. This trend is consistent across all domains, including particularly challenging ones like Creative, CAD and OS, confirming the robustness of our RFT pipeline in complex real-world scenarios.

\begin{figure*}[!h]
    \centering
    \includegraphics[width=1\textwidth]{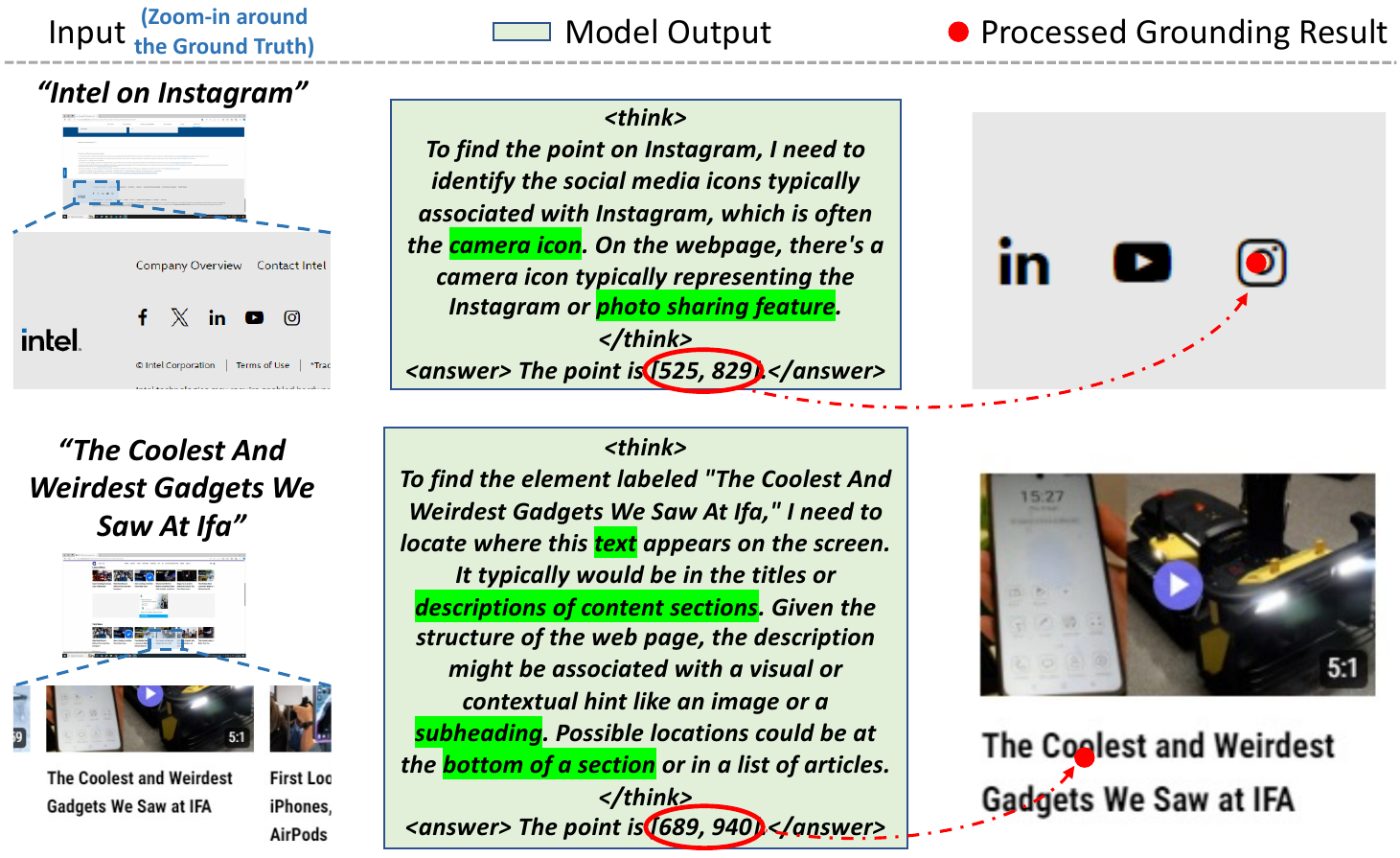}
    \caption{Qualitative Result of \textit{GuirlVG}.}
    \label{qualitative}
\end{figure*}
\section{Qualitative Results}

In this section, we present qualitative results to illustrate the reasoning capabilities of \textit{GuirlVG} in GUI visual grounding tasks. \cref{qualitative} shows two representative examples, each consisting of the input (left), model output with intermediate thinking steps (middle), and the final grounding result (right). The thinking process is highlighted with green color.
In the first example, the task is to locate an icon on a webpage. \textit{GuirlVG} begins by reasoning that it needs to identify social media icons, often represented by a camera icon. Recognizing the webpage context, the model correctly identifies the Instagram icon and grounds the instruction to the coordinates, as shown in the red dot in the grounding result.
In the second example, the instruction is to find a text. \textit{GuirlVG} reasons that the target is a text and it is likely to appear as a contextual hint like a subheading. By analyzing the structure of the webpage, the model further reasons that the target is at the bottom of a section. 
These qualitative results underscore \textit{GuirlVG}'s textual understanding and advanced reasoning abilities, enabled by our reinforcement learning-based approach. By explicitly modeling the thinking process, \textit{GuirlVG} not only achieves high accuracy but also provides interpretable steps, making it a reliable solution for GUI-VG tasks.

%% file: sections/con.tex
\section{Conclusion}

In this work, we revisit the paradigm of post-training for GUI visual grounding and present the first comprehensive empirical study of rule-based reinforcement fine-tuning (RFT) in this domain. Through systematic analysis and a series of targeted innovations—including the decomposition of GRPO components, introduction of the Adversarial KL Factor, and extensive tuning of training configurations—we demonstrate that RFT, when properly optimized, decisively outperforms supervised fine-tuning (SFT). Using as few as 2K training examples, our method surpasses strong SFT baselines trained on orders of magnitude more data across three challenging benchmarks, achieving new state-of-the-art performance. These findings challenge the prevailing reliance on large-scale SFT and highlight RFT as a more data-efficient and generalizable alternative for GUI-VG.

%% file: sections/append.tex
\newpage
\appendix


\section{Additional details for \cref{rft_vs_sft}}\label{appendix_rft_vs_sft}
We provide additional details of the trivial adoption of RFT (RFT-trivial) and the implementation of SFT, which contributes to the reproducibility of the results of this paper.

\begin{algorithm}[h] 
\caption{Format Reward Calculation} 
\label{alg:format-reward} 
\begin{algorithmic}[1] 
\Function{FormatReward}{completion} 
\State $pattern \gets$ regex \verb|"<think>.*?</think>\s*<answer>.*?</answer>"|
\State \Return 1.0 if completion matches $pattern$ else 0.0
\EndFunction 
\end{algorithmic} 
\end{algorithm}

\begin{algorithm}[h]
\caption{Accuracy Reward Calculation}
\label{alg:acc-reward-single}
\begin{algorithmic}[1]
\Function{AccuracyReward}{completion, GT\_box}
\State $answer\_pattern \gets$ regex \verb|<answer>(.*?)</answer>|
\State $bbox\_pattern \gets$ regex \verb|[(\d+),\s*(\d+),\s*(\d+),\s*(\d+)]|
\State $reward \gets 0.0$
\If{completion matches  $answer\_pattern$}
\State $pred\_bbox \gets$ find the match in completion
\If{length of $pred\_bbox$ is 4 \textbf{and} IoU($pred\_bbox$, GT\_box) > 0.5}
\State $reward \gets 1.0$
\EndIf
\EndIf
\State \Return $reward$
\EndFunction
\end{algorithmic}
\end{algorithm}
The format reward function of RFT-trivial is shown in \cref{alg:format-reward} and the corresponding accuracy reward function is shown in \cref{alg:acc-reward-single}.

For the SFT baseline, we use the following prompt:

\textit{Please provide the bounding box coordinates of the region described by this sentence: <description>.}

The answer format is:

\textit{json} \textbackslash n \textit{[{bbox\_2d: <ground-truth bounding box>, label: <description>}]} \textbackslash n.

We adopt the official evaluation code of Qwen2.5-VL\footnote{\url{https://github.com/QwenLM/Qwen2.5-VL/blob/main/cookbooks/computer_use.ipynb}} to obtain the zero-shot baseline performance.

\section{Additional details for \cref{reward_grpo}}\label{appendix_reward_grpo}
We provide details of our Soft Format Reward in \cref{alg:soft-format-reward} to help readers better understand it.

\begin{algorithm}[h] 
\caption{Soft Format Reward Calculation} 
\label{alg:soft-format-reward} 
\begin{algorithmic}[1] 
\Function{SoftFormatReward}{completion} 
\State $score \gets 0$ 
\If{<think>'' in completion} 
\State $score \gets score + 0.5$ 
\EndIf 
\If{</think>'' in completion} 
\State $score \gets score + 0.5$ 
\EndIf 
\If{full <answer>...</answer>'' block detected} 
\State $score \gets score + 2/3$ 
\If{exactly two numbers found inside the block} 
\State $score \gets score + 1/3$ 
\EndIf 
\ElsIf{<answer>'' or ``</answer>'' detected} 
\State $score \gets score + 1/3$ 
\EndIf 
\State 
\Return $score / 1.5$ \Comment{normalized by the maximum possible reward} 
\EndFunction 
\end{algorithmic} 
\end{algorithm}

The prompt we use in \cref{reward_grpo} is as follow:

\textit{Please provide the bounding box coordinates [x1, y1, x2, y2] of a specific element based on this sentence: <description>. First, think about the reasoning process in the mind within <think> </think> tags. Then, output the bounding box coordinates within <answer> </answer> tags.}

\section{Limitations}

While our empirical study on \textit{GuirlVG} is comprehensive, certain limitations remain. First, due to the scope of this work, we focused on a specific multimodal large language model, Qwen2.5-VL~\cite{qwen2.5}, and did not extend our experiments to other models, such as Qwen2-VL~\cite{qwen2}. Including such results could provide a more complete understanding of \textit{GuirlVG}'s generalizability across different MLLM architectures. Second, computational resource constraints prevented us from exploring larger-scale models, such as those with 32B or 72B parameters. Investigating these models could offer insights into the model scale of our approach. Finally, on the data front, while we achieved strong results with limited training samples (2K to 5.2K), access to additional computational resources could enable further exploration of alternative datasets to replicate and validate our performance. These directions, though currently constrained, present valuable opportunities for future work.